\documentclass[letterpaper, 10 pt, conference]{ieeeconf}

\IEEEoverridecommandlockouts
\usepackage{times}
\usepackage{amsmath,amssymb}
\usepackage{graphicx}
\usepackage{booktabs}
\usepackage{multirow}
\usepackage{hyperref}
\usepackage{float}

\usepackage[font=small]{caption}   
\title{\LARGE \bf
Confidence-Gated Robot Autonomy: \\
When Does Uncertainty Actually Help?
}
\author{Johannes A.~Gaus$^{1}$, Jhon P.F. Charaja$^{1}$, and Daniel Haeufle$^{1}$%
\thanks{$^{1}$ Hertie Institute for Clinical Brain Research \& Center for Integrative Neuroscience,
University of Tübingen, Germany}
}

\begin{document}
\bstctlcite{IEEEexample:BSTcontrol}
\maketitle
\thispagestyle{empty}
\pagestyle{empty}

\begin{abstract}
Robotic systems often use predictive uncertainty to decide whether to act autonomously or defer to a fallback policy. In threshold-gated autonomy, uncertainty matters mainly through its ability to rank likely errors. Standard metrics such as expected calibration error and AUROC do not directly test whether uncertainty changes act/defer decisions. We therefore evaluate uncertainty using Spearman rank correlation, paired bootstrap equivalence testing, and act/defer agreement. Across three temporal activity-recognition benchmarks, we find a dataset-dependent competence regime below which uncertainty provides a weak and unstable error ranking. Above this regime, softmax heuristics, MC Dropout, and ensembles produce similar gating behavior, while threshold choice has a much larger effect on execution outcomes. A multi-seed embodied simulation shows the same pattern for collision rate and cost once realized autonomy is matched. Under temporal covariate shift, ranking quality remains stable, but fine-grained semantic OOD detection remains near chance. These results suggest that simple uncertainty proxies can suffice for selective gating once the base model is competent, but not for semantic novelty detection.
\end{abstract}

\section{Introduction}

Robotic systems must estimate not only the most likely state of the world, but also whether that estimate is reliable enough for autonomous execution~\cite{Conlon2024EventTriggered}. A common mechanism is confidence-gated autonomy: a perception model predicts an action or activity label, and an uncertainty score determines whether the robot acts autonomously or defers to a fallback policy. This is closely related to selective prediction and reject-option decision making, which formalize the risk--coverage trade-off induced by abstention~\cite{Geifman2017,Chow1970rejectTradeoff}.

In practice, uncertainty is often derived from softmax probabilities using entropy, margin, or least confidence, although Bayesian approximations and ensembles are also common. This raises a practical question: how should uncertainty be evaluated when its role is to gate execution rather than provide calibrated probabilities? Standard metrics such as expected calibration error (ECE) and area under the receiver operating characteristic curve (AUROC) do not directly answer whether two estimators induce materially different act/defer decisions.

\begin{figure}[t]
    \centering
    \includegraphics[scale=0.11]{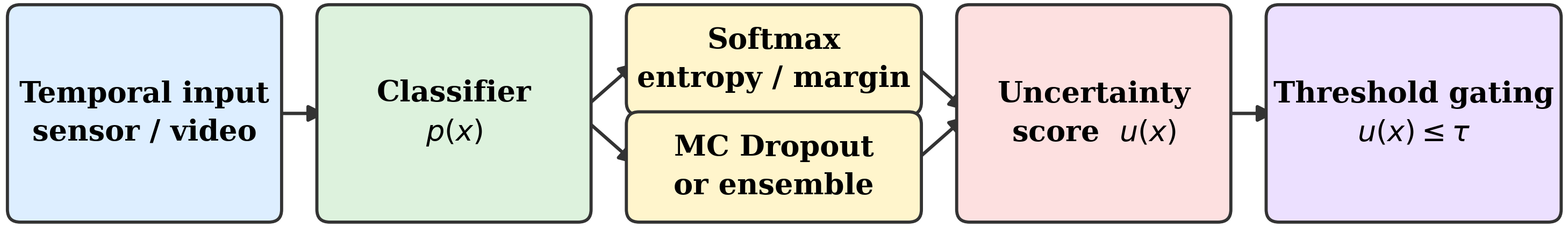}
    \caption{Confidence-gated autonomy: a classifier produces an uncertainty score $u(x)$, and threshold $\tau$ determines whether the robot executes autonomously or defers to fallback. In our experiments, varying $\tau$ changes behavior far more than switching uncertainty estimators.}
    \label{fig:summary_overview}
\end{figure}

We study this question in threshold-gated temporal activity recognition along two axes: model competence and distribution shift. Across three benchmarks, we find that uncertainty becomes decision-useful only after the base model enters a dataset-dependent competence regime. Above this regime, softmax heuristics, MC Dropout, and ensembles induce similar gating behavior, while threshold choice has a much larger effect on execution outcomes. Under temporal covariate shift, ranking quality remains stable enough for selective deferral, whereas fine-grained semantic OOD detection remains weak. Figure~\ref{fig:summary_overview} summarizes the setting.

\textbf{Contributions.}
\begin{itemize}
    \item We introduce a decision-centric evaluation protocol for confidence-gated autonomy based on rank correlation, paired bootstrap equivalence testing, and act/defer agreement.
    \item We identify dataset-dependent competence regimes below which uncertainty does not reliably rank prediction errors for gating.
    \item We show that threshold choice has a substantially larger effect on execution outcomes than estimator choice, both offline and in a minimal embodied simulation, while fine-grained semantic OOD detection remains near chance.
\end{itemize}

\section{Related Work}
\label{sec:related_work}

\textbf{Uncertainty Quantification and Calibration.} 
Predictive uncertainty methods include MC Dropout~\cite{GalGhahramani2016Dropout} and deep ensembles~\cite{Lakshminarayanan2017DeepEns,Ovadia2019PredUncert}. However, improved calibration does not necessarily improve error ranking for selective prediction~\cite{Ding2020ReviseUncert}.

\textbf{Autonomy Adjustment and Risk-Aware Robotics.}
In robotics, uncertainty is used for selective execution, shared autonomy, and introspection~\cite{Nava2021SPatial,Conlon2024EventTriggered,Javdani2018Hindsight,Jonnavittula2025SARI}, as well as risk-sensitive control through chance constraints or CVaR~\cite{ORIGANE2025348}. Our focus is narrower: when uncertainty is useful for threshold-gated autonomy and when common proxies are effectively interchangeable.

\textbf{Task-Driven Shift Detection.}
Related work detects distribution shift through task-level performance violations rather than input novelty alone~\cite{Farid2024TaskDriven}. This helps explain why temporal covariate shift and semantic held-out-class detection behave differently in our experiments.

\textbf{Selective Prediction and Evaluation Pitfalls.} 
Selective prediction formalizes the coverage--risk trade-off~\cite{Chow1970rejectTradeoff,Geifman2017}. Common evaluations use risk--coverage curves, SelectiveNet, or AUROC~\cite{Geifman2019SelectiveNetAD,Cortes2016Reject}, but global metrics such as ECE or AUROC may fail to reflect decision-level utility in structured robotic tasks~\cite{Ding2020ReviseUncert}. Recent work has also highlighted pitfalls in multi-threshold evaluation of selective systems~\cite{Traub2024CommonFlaws}. In robotics, related conformal and risk-calibrated methods focus on safety guarantees or action-set calibration rather than whether scalar uncertainty scores induce different threshold-gated act/defer behavior~\cite{Lindemann2023Safe,Lidard2024RiskCalibrated}.

\textbf{Competence-Dependent Utility of Uncertainty.} 
Aleatoric and epistemic uncertainty remain central in uncertainty quantification~\cite{KIUREGHIAN2009105,KendallGal2017Uncert,Depeweg2018Decomp}. Our focus is complementary: when increasing base-model accuracy makes uncertainty reliable enough to rank prediction errors for gating.

\section{Problem Formulation: Confidence-Gated Autonomy}

We consider a temporal classifier that maps each input sequence $x \in \mathcal{X}$ to a predictive distribution over labels $\mathcal{Y}$,
\begin{equation}
    \mathbf{p}(x) = [p_1(x), \dots, p_K(x)], \qquad \sum_{k=1}^{K} p_k(x) = 1,
\end{equation}
with prediction $\hat{y}(x)=\arg\max_k p_k(x)$. Each input is also assigned a scalar uncertainty score $u(x)$, used only for ranking, with larger values indicating greater uncertainty. Confidence-like scores such as margin are sign-reversed so that all methods share the same ordering convention.

Autonomous execution is controlled by threshold $\tau$:
\begin{equation}
    \pi_\tau(x)=
    \begin{cases}
        1, & u(x)\le\tau,\\
        0, & \text{otherwise},
    \end{cases}
\end{equation}
where $\pi_\tau(x)=1$ denotes autonomous execution and $\pi_\tau(x)=0$ deferral. Smaller $\tau$ lowers coverage and executed error; larger $\tau$ increases both.

\section{A Decision-Centric Evaluation Framework}
We evaluate uncertainty as a decision variable rather than a probabilistic quantity. The framework combines Spearman rank correlation, paired bootstrap equivalence testing, and act/defer agreement under threshold sweeps to test whether differences between methods are operationally meaningful.

\subsection{Uncertainty Scores}
We evaluate seven scores: four softmax-derived heuristics (entropy, margin, variation ratio, least confidence), MC Dropout predictive entropy, MC Dropout epistemic variance from $T=30$ stochastic passes, and 3-member ensemble entropy. These span lightweight deterministic proxies and standard posterior approximations~\cite{GalGhahramani2016Dropout,Lakshminarayanan2017DeepEns}.

\subsection{Metrics for Ranking-Based Gating}

\textbf{Ranking Quality.}
Because confidence-gated execution depends primarily on the ordering of predictions by uncertainty, we evaluate ranking quality using Spearman rank correlation between uncertainty scores and the binary error indicator,

\begin{equation}
\rho = \mathrm{corr}_{\mathrm{rank}}\!\left(u(x), \mathbb{1}\!\left[\hat{y}(x)\neq y\right]\right).
\end{equation}
Here, $y$ denotes the ground-truth label. Although the error indicator is binary, Spearman $\rho$ still captures whether higher uncertainty is assigned to erroneous predictions more often than to correct ones, which is the key requirement for confidence-gated autonomy.

\textbf{AUROC.}
We report AUROC as a familiar baseline, but our main conclusions rely on rank correlation, equivalence testing, and decision-level agreement because these align more directly with threshold-gated autonomy.

\subsection{Paired Bootstrap and Practical Equivalence}

All metrics are estimated with nonparametric bootstrap resampling over test examples ($B=1000$). When comparing methods, we use paired bootstrap resampling so that metric differences are computed on the same resampled test sets. Rather than testing only for nonzero differences, we assess practical equivalence with margin $\delta = 0.01$ for ranking metrics~\cite{Lakens2017Equivalence}. We use $\delta = 0.01$ as a conservative small-effect threshold because differences of this magnitude in Spearman $\rho$ were not expected to induce materially different coverage-matched act/defer decisions in our setting. Specifically, two methods are considered operationally equivalent if the bootstrap 95\% confidence interval (CI) for their metric difference lies entirely within $[-\delta, +\delta]$. For decision-level comparisons, thresholds are matched by coverage rather than by raw score value, since uncertainty scores can live on different scales across estimators.






\subsection{Covariate Shift Protocol}

To evaluate open-world robustness, we introduce temporal corruptions to test sets at five severity levels. For each input sequence, we apply one of three corruption types: (1) temporal frame dropout (drop 10\%, 20\%, \ldots, 50\% of frames randomly), (2) Gaussian noise (add i.i.d.\ noise with $\sigma = 0.05, 0.10, 0.20, 0.30, 0.50$), or (3) temporal frame jitter (randomly shuffle 10\%, 20\%, \ldots, 50\% of frame pairs). For each corruption and severity, we measure: accuracy on corrupted data, mean uncertainty (entropy), execution error at a fixed deferral threshold ($\tau^*$ calibrated for 80\% coverage on clean data), and Spearman $\rho$ between uncertainty and errors. This protocol directly measures whether confidence-gated autonomy degrades gracefully or fails abruptly under distribution shift.

\section{Experiments and Results}

We evaluate on three temporal activity-recognition benchmarks spanning distinct accuracy regimes:
\begin{itemize}
    \item \textbf{EGO4D:} Egocentric action recognition, 7 classes, 1875 test samples. Test accuracy $\approx 63.5\%$.
    \item \textbf{iAssist:} Assistive activity recognition from wearable IMU and gaze during daily living, 8 classes, 6514 test samples. Test accuracy $\approx 96.5\%$.
    \item \textbf{GTEA:} Gaze-plus activity recognition, 20 classes, 1954 test samples. Test accuracy $\approx 46\%$.
\end{itemize}


Our base classifier is a fixed hybrid multimodal temporal architecture with modality-specific encoders, Perceiver IO fusion, and prediction heads. All models are trained with cross-entropy and frozen during uncertainty evaluation. Results are aggregated over three random seeds unless noted otherwise. Uncertainty is computed from a single forward pass for softmax-based scores, $T=30$ stochastic passes for MC Dropout, and three ensemble members; all metrics use $B=1000$ paired bootstrap resamples.

\subsection{Competence Regime: When Uncertainty Works}

\begin{figure}[t]
    \centering
    \includegraphics[width=\columnwidth]{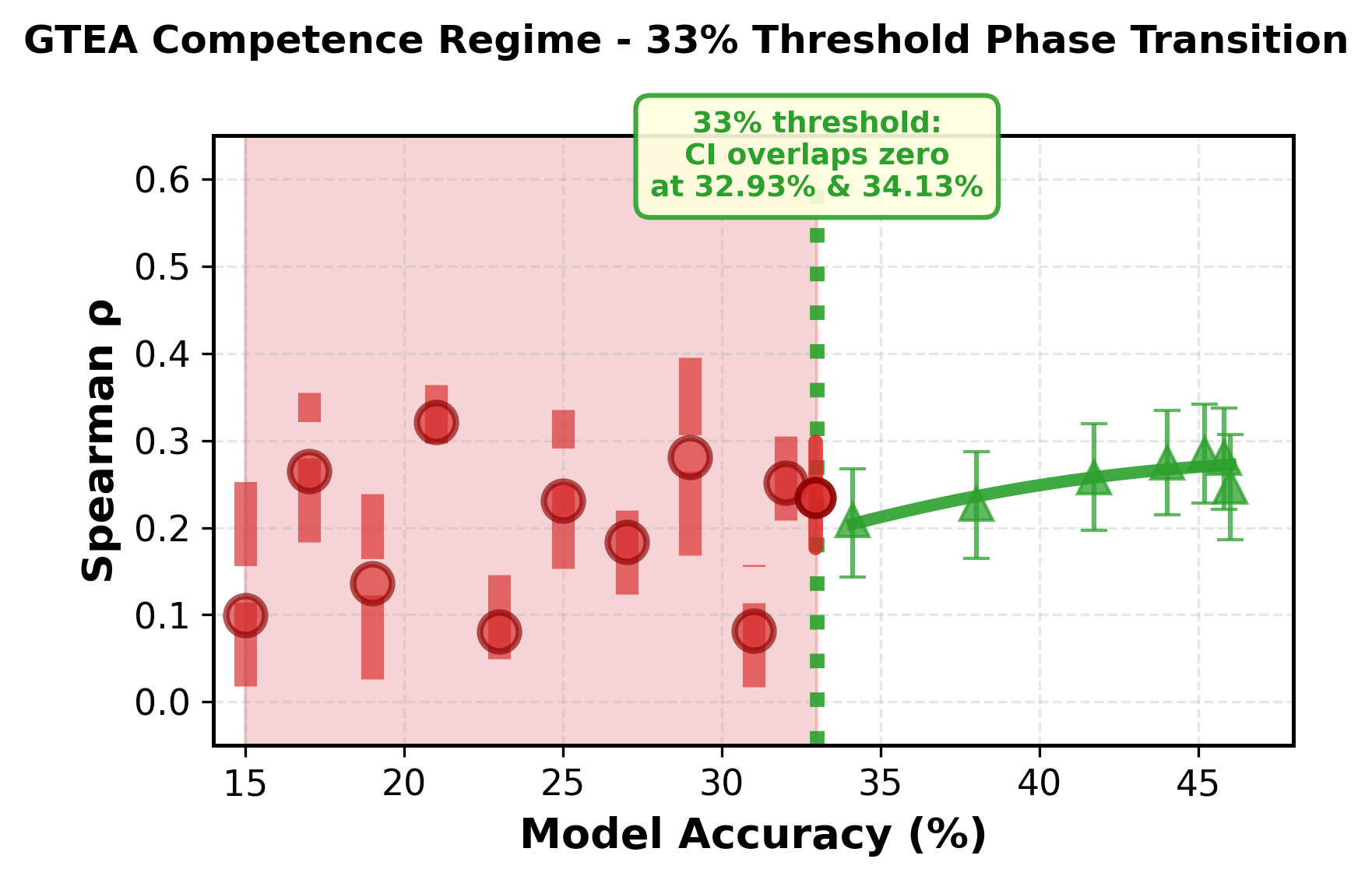}
    \caption{Competence transition on GTEA. Below the threshold, Spearman $\rho$ is weak and variable across runs; above it, ranking becomes more stable. Similar transitions appear on EGO4D and iAssist.}
    \label{fig:competence}
\end{figure}

Uncertainty becomes useful for gating only when the model is accurate enough for its errors to be ranked reliably by confidence. We call a model \emph{competent for gating} when Spearman $\rho$ is consistently positive and stable across runs (Fig.~\ref{fig:competence}).

\textbf{Identifying competence thresholds.}
To probe different competence levels, we train the same architecture on 5\%, 10\%, 20\%, 40\%, and 70\% subsets of the training data, using three random seeds per subset. This yields a range of base-model accuracies for each dataset. We define the competence transition as the lowest accuracy level at which, across all three seeds, the lower bound of the 95\% bootstrap confidence interval for Spearman $\rho$ exceeds zero and the inter-seed standard deviation of $\rho$ falls below $0.05$. Essentially, we define a model as ``competent'' when its uncertainty scores provide a consistent and positive signal for ranking errors across different training runs. We use this as an operational criterion for the onset of decision-useful uncertainty; the resulting thresholds are dataset-specific operating markers rather than universal constants.

Across all three datasets, the absolute threshold differs, but the qualitative pattern is consistent (Fig.~\ref{fig:competence}). On GTEA, the transition occurs around 33\% accuracy; on EGO4D, around 58\%; and on iAssist, around 70\%. Below these regimes, uncertainty provides a weak ranking signal with higher variability across seeds. Above them, uncertainty yields a more stable ordering of likely errors.

\subsection{Method Equivalence: Softmax, MC Dropout, and Ensembles}

Within competent regimes, we evaluate 7 estimators: softmax entropy, softmax margin, softmax variation ratio, softmax least confidence, MC Dropout entropy, MC Dropout epistemic variance, and ensemble entropy (3-member).

\begin{figure}[t]
    \centering
    \includegraphics[width=\columnwidth]{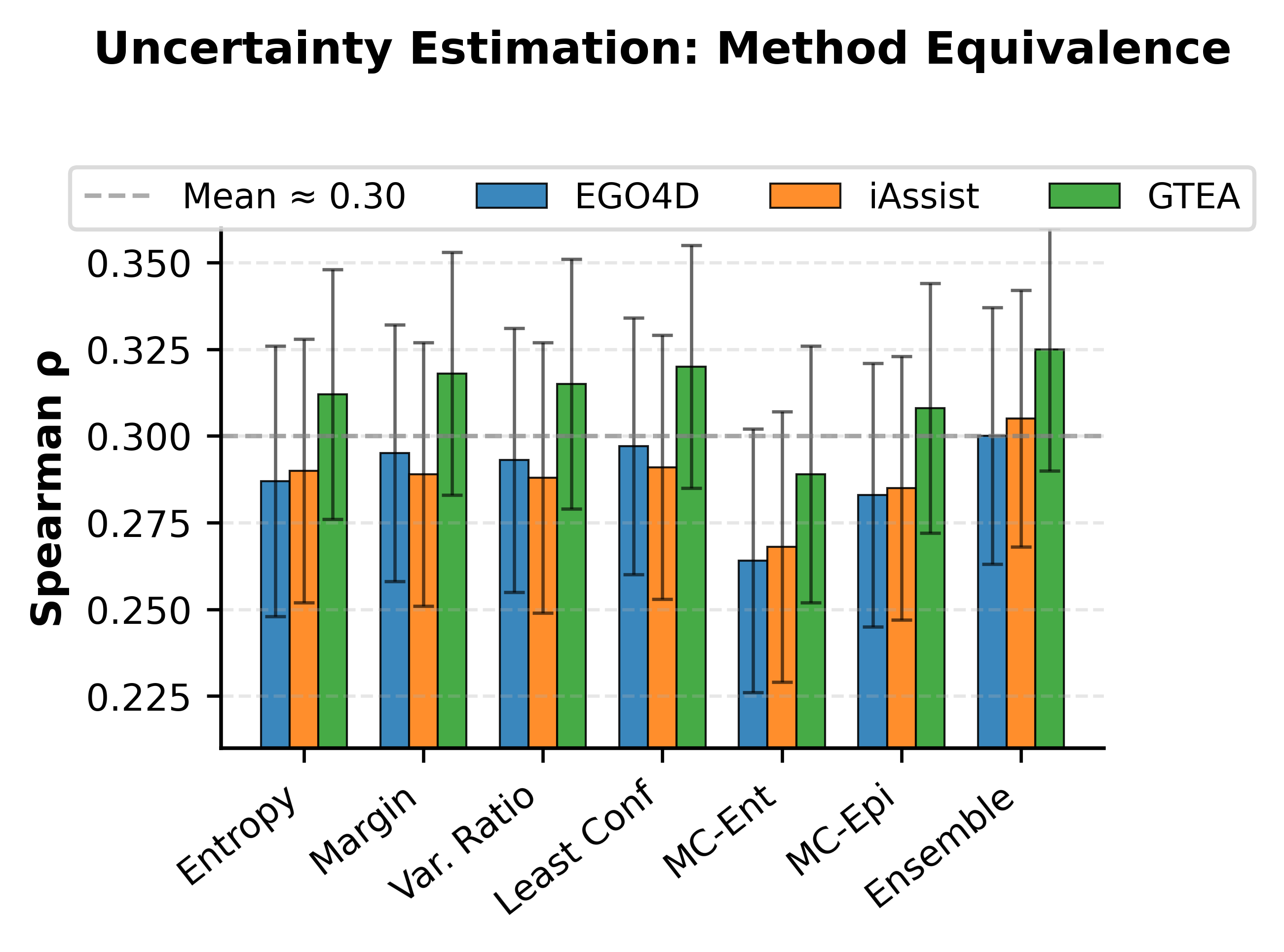}
    \caption{Estimator comparison across datasets. All seven methods achieve similar Spearman $\rho$ with strongly overlapping bootstrap intervals; remaining metric differences do not translate into materially different gating behavior.}
    \label{fig:methods}
\end{figure}

\textbf{Softmax Heuristics.}
On EGO4D (63.5\% accuracy), all softmax-derived heuristics achieve Spearman $\rho$ in the range $[0.287, 0.288]$ with bootstrap CI widths of approximately $0.036$. Paired equivalence testing shows all pairwise differences lie within the $\delta = \pm 0.01$ band. Figure~\ref{fig:methods} shows that all estimators cluster tightly around $\rho \approx 0.30$ with strongly overlapping confidence intervals. Among the softmax heuristics, pairwise metric differences are $<10^{-3}$.
Entropy-based scores are normalized to $[0,1]$ when needed so that threshold values are comparable and interpretable across datasets with different numbers of classes.

\textbf{MC Dropout and Ensembles.}
MC Dropout entropy achieves $\rho \approx 0.264$ and epistemic variance $\rho \approx 0.283$, whereas ensemble entropy reaches $\rho \approx 0.300$. MC Dropout entropy falls outside the $\delta = 0.01$ metric equivalence band ($\Delta\rho = -0.023$), yet act/defer agreement remains high across the full threshold sweep (minimum agreement: $97.8\%$). Thus, the metric difference does not translate into materially different robot-level decisions. For gating, this behavioral validation is the more relevant criterion: after matching thresholds by coverage, all seven estimators produce nearly indistinguishable act/defer sets, despite modest differences in global ranking metrics (Table~\ref{tab:estimator_spearman}).


\begin{table}[t]
\centering
\caption{Spearman $\rho$, AUROC, and act/defer agreement relative to softmax entropy for 7 uncertainty estimators on EGO4D (63.5\% accuracy). All estimators achieve $\rho \approx 0.27$--$0.30$ with high behavioral agreement.}
\label{tab:estimator_spearman}
\begin{tabular}{lccc}
\toprule
Estimator & Spearman $\rho$ & AUROC & Agr.\ vs.\ ent.\ (\%) \\
\midrule
Softmax Entropy & $0.288$ & $0.666$ & -- \\
Softmax Margin & $0.288$ & $0.666$ & $97.9$ \\
Softmax Var.\ Ratio & $0.287$ & $0.666$ & $97.8$ \\
Softmax L.C.\! & $0.287$ & $0.666$ & $97.9$ \\
MC Dropout Entropy & $0.264$ & $0.647$ & $98.1$ \\
MC Dropout Epistem. & $0.283$ & $0.660$ & $99.2$ \\
Ensemble Entropy & $0.300$ & $0.673$ & $99.5$ \\
\bottomrule
\end{tabular}
\end{table}

\subsection{Threshold Choice Has Larger Behavioral Effect Than Estimator Choice}

Sweeping the execution threshold $\tau$ changes the conservativeness of the autonomy policy directly. On EGO4D, restricting execution to the 10\% most confident predictions reduces execution error from 36.5\% to 19.9\%, a 16.6 percentage-point or 46\% relative reduction. By comparison, replacing softmax entropy with softmax margin changes execution accuracy by less than 0.5\%.

Across coverage-matched threshold sweeps, all seven estimators produce nearly identical coverage--risk behavior, with act/defer agreement typically above $99\%$. In this setting, threshold selection therefore has a substantially larger practical effect on execution outcomes than the choice of uncertainty estimator.

\subsection{Embodied Consequences in a Minimal Navigation Simulation}

To test whether offline gating differences survive contact with a downstream robot decision layer, we evaluated a minimal embodied simulation in PyBullet using EGO4D, which lies above its competence threshold while remaining far from ceiling accuracy. The simulator defines a stylized navigation consequence model in which the robot must choose one of three maneuver primitives (left, straight, right) to pass through a constrained passage. For embodied EGO4D simulation, the 7-way activity label is deterministically projected to 3 navigation primitives via $m(y)=y \bmod 3$ (left: $\{0,3,6\}$, straight: $\{1,4\}$, right: $\{2,5\}$); the same mapping is applied to ground truth $m(y)$ and prediction $m(\hat{y})$, so simulation correctness is defined at the mapped primitive level. The uncertainty score then determines whether the robot executes the mapped maneuver or defers to a fallback policy. Incorrect autonomous maneuvers can produce collisions, whereas deferred episodes execute a conservative fallback with fixed cost. We report results over three independently trained model seeds and 20 stochastic scene repeats per seed.

\begin{figure}[t]
    \centering
    \includegraphics[width=\columnwidth]{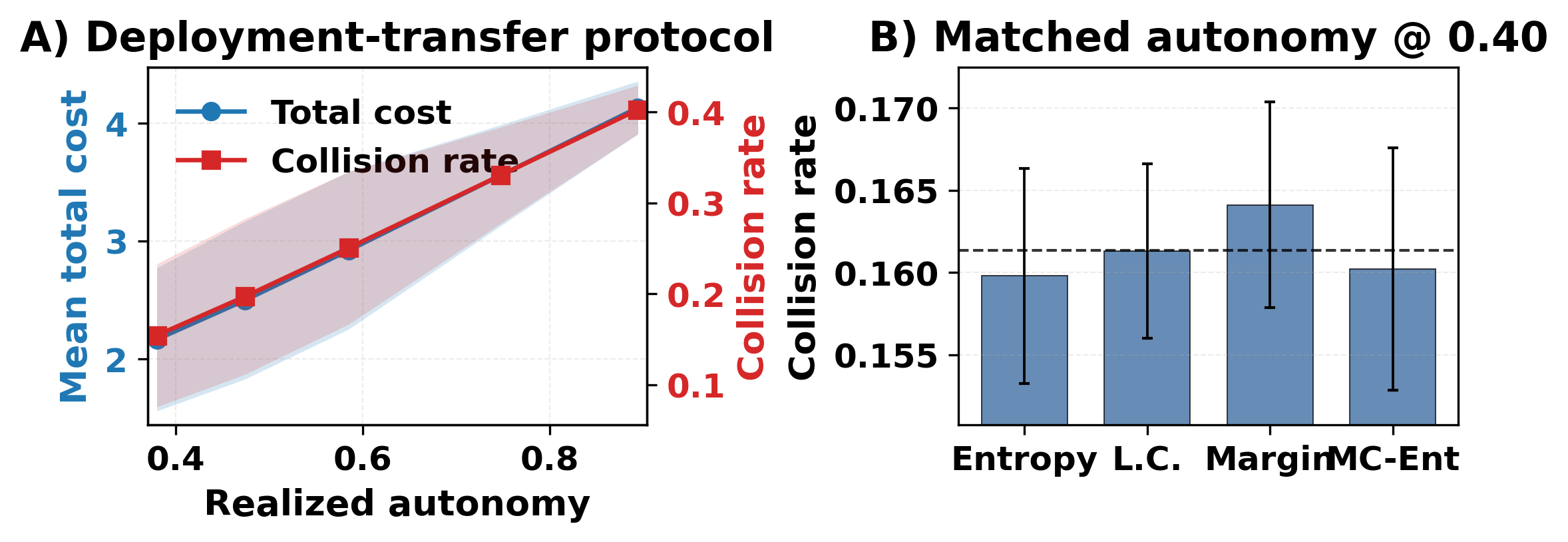}
    \caption{Embodied simulation results on EGO4D. (A) Deployment-transfer protocol using entropy: increasing realized autonomy strongly increases both mean total cost (blue) and collision rate (red), showing that operating-point choice dominates embodied outcomes. Shaded regions indicate variability across seeds. (B) Matched realized autonomy ($0.40$): entropy, least confidence (L.C.), margin, and MC Dropout entropy (MC-Ent) yield very similar collision rates, with error bars denoting seed standard deviation. The dashed line indicates the mean across methods.}
    \label{fig:simulation}
\end{figure}

This setup is not intended to model a full manipulation or navigation stack; rather, it serves to test whether uncertainty-induced act/defer differences that appear offline remain materially different once mapped to simple downstream robot consequences.

We evaluate two protocols: \emph{deployment-transfer} (thresholds calibrated for coverages $\{0.5,\ldots,0.9\}$, transferred to simulation) and \emph{matched-realized-autonomy} (estimators compared at identical realized autonomy, isolating estimator choice from calibration drift).

The deployment-transfer results confirm that operating-point choice dominates: total cost rises from $1.52$ to $3.73$ and collision rate from $0.072$ to $0.352$ across the autonomy range (Fig.~\ref{fig:simulation}, left). At matched realized autonomy (${\approx}0.40$), all four primary estimators produce nearly identical consequences (Fig.~\ref{fig:simulation}, right).

We also tested sensitivity to key deployment assumptions. Across the examined collision-to-deferral cost ratios, the relative ordering among the main estimators remained qualitatively similar, suggesting that the near-equivalence result is not tied to a single cost choice. By contrast, replacing the safe-wait fallback with straight motion substantially increased absolute cost and collision, showing that fallback design strongly affects consequence magnitude even when the relative estimator comparison changes little. We therefore interpret the simulation as consequence-level validation of the main decision-centric claim rather than as a substitute for evaluation on a physical robot or in richer long-horizon control settings.

\subsection{Cost-Sensitive Threshold Optimization}
Once uncertainty is decision-useful, the remaining deployment question is how to choose the operating threshold. Given a validation set and application-specific costs, practitioners can select $\tau$ by minimizing expected decision cost:

\begin{equation}
J(\tau) = C_{\mathrm{err}} \cdot \mathrm{Risk}(\tau) \cdot \mathrm{Cov}(\tau) + C_{\mathrm{def}} \cdot (1 - \mathrm{Cov}(\tau)),
\end{equation}

where $\mathrm{Cov}(\tau)$ is coverage and $\mathrm{Risk}(\tau)$ is the executed error rate at threshold $\tau$. This objective assumes a fixed cost for deferral and does not model downstream failures of the fallback policy; it is intended as a simple deployment-level decision model rather than a full control-theoretic cost functional.

Sweeping cost ratios $C_{\mathrm{err}} / C_{\mathrm{def}} \in \{0.5, 1, 2, 5, 10, 20\}$, the optimal threshold $\tau^*$ ranges from 1.0 (accept all predictions) at low cost ratios to 0.236 (high deferral) at cost ratio 20. This makes the deployment trade-off explicit. The cost parameters $C_{\mathrm{err}}$ and $C_{\mathrm{def}}$ must be specified by the practitioner; the framework does not estimate them. This also helps explain why $\rho \approx 0.30$ can still be practically useful: for threshold-gated autonomy, the relevant question is not perfect global ranking, but whether the score supports an acceptable risk--coverage operating point under the chosen cost ratio.

\section{Open-World Robustness: Temporal Covariate Shift and Semantic OOD}

The analyses above consider clean, closed-set test data. Real deployments, however, can deviate from this setting in at least two qualitatively different ways. First, inputs may remain within the known label space but become corrupted or temporally degraded, requiring uncertainty to continue ranking which in-distribution predictions are reliable enough for autonomous execution. Second, genuinely novel semantic classes may appear at test time, requiring uncertainty to signal that the input does not belong to the known class set. We therefore evaluate robustness under both settings: controlled temporal covariate shift and semantic out-of-distribution (OOD) detection via held-out classes.

\subsection{Graceful Degradation Under Temporal Covariate Shift}

Applying temporal corruptions to test inputs, we measure whether uncertainty estimates degrade gracefully or fail abruptly. Each corruption is evaluated at five severity levels, from mild (severity 1) to strongest (severity 5). To make the overall behavior easy to interpret, the dataset-specific discussion below focuses on severity 5 as the most demanding setting.

\textbf{EGO4D Results.}
Under temporal frame dropout at severity 5 (50\% of frames removed), accuracy remains largely stable ($\approx 60.8\%$), while mean uncertainty increases from 0.803 to 0.813. Using the clean-data threshold $\tau^* = 0.476$ (80\% coverage on clean data), execution error drops from 38.8\% to 11.5\%, a 3.4$\times$ reduction. Spearman $\rho$ remains stable at 0.302, indicating that uncertainty ranking quality is preserved under corruption.

Gaussian noise and temporal jitter, by contrast, cause only minimal uncertainty changes ($<0.8\%$), suggesting that the model is more sensitive to disrupted temporal structure than to low-level perturbations.

\textbf{iAssist Results.}
Temporal frame dropout produces a larger increase in uncertainty (0.039 to 0.059) and a modest drop in accuracy (96.3\% to 93.5\%). Using the clean-data threshold $\tau^* = 0.014$ (80\% coverage on clean data), execution error drops from 3.47\% to 0.10\%. While this corresponds to a 36.4$\times$ relative reduction, the large ratio partly reflects the low baseline error; the absolute improvement is 3.37 percentage points. Spearman $\rho$ reaches 0.368, indicating especially strong error ranking under temporal dropout in this dataset.

\textbf{GTEA Results.}
GTEA (46\% baseline accuracy) operates near the competence boundary and exhibits minimal error reduction via gating. Under temporal frame dropout at severity 5, execution error remains at 0.560, identical to the baseline error. This absence of improvement reflects the model's insufficient confidence margins for effective deferral when operating near the competence threshold. Spearman $\rho$ remains stable at 0.278, confirming that ranking quality is preserved even when the model lacks sufficient accuracy to translate that ranking into useful gating decisions.

\textbf{Interpretation.}
EGO4D and iAssist exhibit graceful degradation: gating reduces execution error while Spearman $\rho$ remains broadly stable (mean $0.296$ across datasets and corruption types). GTEA, near the competence boundary, shows little benefit, consistent with the competence-regime finding. The selective response to temporal coherence loss, but not to Gaussian noise or temporal jitter, is consistent with uncertainty reacting more strongly to semantically disruptive temporal degradation than to lower-level perturbations.


\subsection{Fine-Grained Semantic OOD Detection Remains Near Chance}

In contrast to the temporal covariate-shift results above, fine-grained semantic OOD detection remains weak when evaluated using randomly held-out action classes.

\textbf{Standard Held-Out Classes.}
Using 20\% randomly held-out classes as OOD, we compute AUROC for uncertainty-based detection.

On EGO4D, softmax entropy achieves AUROC = 0.35, well below chance. The model is more confident on held-out classes (mean max-prob $0.738$) than on in-distribution classes ($0.663$, $+7.4$ percentage points), likely because held-out actions share visual and motion features with training classes, causing confident assignment to the nearest known category. This is a known failure mode of softmax-based OOD detection~\cite{Hendrycks2016ABF}.

On iAssist, held-out-class detection is near chance (AUROC = 0.49), with held-out classes again slightly more confident (0.989 vs.\ 0.985). On GTEA, AUROC = 0.42, indicating similarly weak class-level OOD detection.

\textbf{Hard OOD Split (Rare Classes).}
Designating the lowest-frequency 20\% of classes as OOD improves detectability on EGO4D (AUROC = 0.67), indicating that semantic class dissimilarity matters. However, this corresponds to a more separated OOD setting and is less representative of the fine-grained in-domain shifts considered above.

\textbf{Why the two shift settings differ.}
Under temporal corruption, the model remains within the known label space, so uncertainty only needs to rank which in-distribution predictions have become less reliable. Held-out-class detection is harder because it requires recognizing semantic novelty. When unseen classes overlap with seen ones in visual or motion structure, softmax-based scores can remain confidently wrong by mapping them to the nearest known class. Thus, uncertainty that is useful for selective prediction may still be inadequate for semantic novelty detection.

\textbf{Implication.}
In our experiments, softmax-based uncertainty is useful for temporal covariate shift but not for fine-grained semantic shift within a domain, motivating dedicated open-set mechanisms as discussed in Section~\ref{sec:discussion}.

\section{Discussion}
\label{sec:discussion}
Our results support a decision-centric view of uncertainty evaluation in robotics. For threshold-gated autonomy, the key question is not only whether uncertainty is calibrated, but whether it changes execution decisions. In our setting, the main practical factor is not estimator choice but whether the base model has reached a competence regime in which its errors can be ranked reliably enough for gating. This transition likely depends not only on overall accuracy but also on error structure, including class overlap, imbalance, and confidence margins. It is also consistent with the asymmetry between temporal covariate shift and semantic novelty detection: ranking likely errors within the known label space is easier than recognizing inputs outside it.

Relative to prior work~\cite{Ding2020ReviseUncert,Geifman2017,Conlon2024EventTriggered,Nava2021SPatial,ORIGANE2025348}, calibration remains important for probabilistic statements, but ranking quality is more relevant when uncertainty gates execution. Once the model is sufficiently competent, MC Dropout and ensembles~\cite{GalGhahramani2016Dropout,Lakshminarayanan2017DeepEns,Ovadia2019PredUncert} offer little practical advantage over softmax heuristics for this threshold-gated setting.

These findings apply most directly to scalar threshold gating and may not extend to settings with asymmetric costs, sequential error compounding, or continuous autonomy modulation~\cite{Javdani2018Hindsight,ORIGANE2025348}. A natural next step is \emph{competence-aware risk-controlled gating}: first test whether uncertainty is decision-useful, then activate a risk-controlled act/defer policy.

An important limitation is that absolute embodied outcomes depend strongly on the fallback policy and on the simplified consequence model used in simulation. The EGO4D simulation also uses a coarse deterministic projection from  7 activity classes to 3 navigation primitives ($m(y) = y \bmod 3$), so it should be interpreted as a controlled consequence model for act/defer decisions rather than a semantics-preserving activity-to-control pipeline; future work should validate with semantically grounded action-to-control mappings. In our setting, replacing a safe-wait fallback with a straight-motion fallback substantially increases cost and collision, even though the relative comparison among the main uncertainty estimators remains similar. The embodied analysis therefore isolates the downstream consequence structure of thresholded act/defer decisions under stylized assumptions rather than estimating full system-level risk. It provides consequence-level grounding for the decision-centric claim, but it does not replace evaluation on a physical robot or in richer long-horizon control settings.

In practice, competence should be established during validation or monitored online only when delayed supervision or reliable proxy feedback is available. When that condition is not met, or when semantic novelty is suspected, the system should revert to conservative fallback or invoke a dedicated open-set mechanism. This links selective autonomy, threshold tuning, and open-world robustness in a single deployment loop, consistent with recent work on conformal safety and risk-calibrated robotic decision making~\cite{Lindemann2023Safe,Lidard2024RiskCalibrated,Farid2024TaskDriven}.

The open-world results also suggest a division of roles: the uncertainty proxies studied here appear adequate for temporal covariate shift via threshold adjustment, but not for fine-grained semantic OOD detection. When semantic shift is expected, dedicated open-set mechanisms remain necessary.

\section{Conclusion}

We studied uncertainty as a decision variable for confidence-gated robot autonomy. Across three temporal activity-recognition benchmarks, uncertainty became useful for gating only after the base model entered a dataset-dependent competence regime. Above this regime, softmax heuristics, MC Dropout, and ensembles produced similar act/defer behavior, whereas threshold choice had a substantially larger effect on execution outcomes. A minimal embodied simulation showed the same qualitative pattern at the consequence level. Under temporal covariate shift, ranking quality remained stable enough for effective deferral, but fine-grained semantic OOD detection remained near chance. These results suggest a practical deployment order: establish task-level competence, tune the execution threshold for the desired risk--coverage trade-off, and add dedicated open-set mechanisms when semantic shift is expected.

\bibliographystyle{IEEEtran}
\bibliography{AllRefs}

\end{document}